\documentclass[12pt]{article} 
\usepackage{amsmath,amsthm,amssymb,bm,mathrsfs}
\usepackage{graphicx,subfigure}
\usepackage{caption,enumerate,listings,setspace}
\usepackage[colorlinks,linkcolor=red,anchorcolor=blue,citecolor=blue]{hyperref}
\usepackage[margin=1in]{geometry}
\usepackage[title]{appendix}
\usepackage{enumitem}
\theoremstyle{plain}
\usepackage{booktabs}
\usepackage[numbers]{natbib}
\usepackage{array}
\usepackage{algorithm}
\usepackage{algpseudocode}
\usepackage{makecell}

\numberwithin{example}{section}

\usepackage[utf8x]{inputenc}
\usepackage{mathrsfs}
\usepackage{amssymb}
\usepackage{amsfonts}
\usepackage{amsmath}
\usepackage{amsthm,verbatim}
\usepackage{wrapfig}
\usepackage{multirow}
\usepackage{longtable}
\usepackage{enumerate}
\usepackage{color, graphicx}
\usepackage{tikz}

\def\boxit#1{\vbox{\hrule\hbox{\vrule\kern6pt
          \vbox{\kern6pt#1\kern6pt}\kern6pt\vrule}\hrule}}

\title{\textbf{
Improve Fidelity and Utility of Synthetic Credit Card Transaction Time Series from Data-centric Perspective}} 

\author
{
Din-Yin Hsieh\thanks{Undergraduate Student, Department of Statistics and Data Science, UCLA, CA, 90095. Email: darrenhsieh1205@g.ucla.edu},
Chi-Hua Wang \thanks{Postdoctoral Scholar, Department of Statistics and Data Science,  UCLA, CA, 90095. Email: chihuawang@ucla.edu},
Guang Cheng\thanks{Professor, Department of Statistics and Data Science, UCLA, CA, 90095. Email: guangcheng@ucla.edu}
}

\begin{document} 

\maketitle

\begin{abstract}
Exploring generative model training for synthetic tabular data, specifically in sequential contexts such as credit card transaction data, presents significant challenges. This paper addresses these challenges, focusing on attaining both high fidelity to actual data and optimal utility for machine learning tasks. We introduce five pre-processing schemas to enhance the training of the Conditional Probabilistic Auto-Regressive Model (CPAR), demonstrating incremental improvements in the synthetic data's fidelity and utility. Upon achieving satisfactory fidelity levels, our attention shifts to training fraud detection models tailored for time-series data, evaluating the utility of the synthetic data. Our findings offer valuable insights and practical guidelines for synthetic data practitioners in the finance sector, transitioning from real to synthetic datasets for training purposes, and illuminating broader methodologies for synthesizing credit card transaction time series.
\end{abstract}

\bigskip
\noindent{\bf Key Words:} Credit Card Transaction Data, Time Series Generative Model, Synthetic Training Datasets, Fraud Detection, Data-centric Machine Learning..

\section{Introduction}
\label{sec:intro}

Synthetic Tabular Data is garnering increasing interest from the academic and industrial sectors alike \cite{jordon2022synthetic}. This surge in attention can be attributed to the capability of synthetic data to foster innovation and aid in decision-making processes, all while adhering to contemporary privacy regulations such as GDPR \cite{EUdataregulations2018} and CCPA \cite{CCPA2018}. The promising prospects of accelerating and supporting the entire machine learning lifecycle make this area particularly appealing. As a result, there is a growing impetus among machine learning researchers and practitioners to delve into the advantages and limitations associated with utilizing synthetic datasets \cite{visani2022enabling, el2020practical}.


While synthetic tabular data offers substantial potential, its application is fraught with challenges, particularly in handling sequential tabular data \cite{Zhang2022SequentialMI}. Despite its apparent similarity to standard tabular data in terms of representation, sequential tabular data is unique due to the existence of \textit{relationships between rows}. These inter-row relationships arise from the specific nature of the application used during data collection. Consider, for example, health monitoring data: here, data points (or rows) are recorded at consistent time intervals, resulting in a regular and predictable pattern conducive to statistical modeling \cite{Martino2022ExplainableAF, Yuan2023EHRDiffER}. In contrast, credit card transaction data presents a different scenario; the time intervals between data points (transactions) are irregular, mirroring the sporadic nature of consumer spending. Consequently, the regularity of data point recording emerges as a crucial consideration in the development of models for synthetic sequential data.

In this study, we aim to offer insights on training generative models tailored for the \textit{credit card transaction data synthesis}, ensuring both high fidelity to real data and optimal utility for machine learning applications. It's important to note that credit card transaction data is inherently multi-sequence, marked by its irregular measurement intervals \cite{Weerakody2021ARO}. Presently, the CPAR (Conditional Probabilistic Auto-Regressive Model) \cite{Zhang2022SequentialMI} stands out in literature as a quintessential multi-sequence generative model. CPAR model is available to researchers and practitioners through the open-source Synthetic Data Vault (SDV) library \cite{Patki2016TheSD} (for a detailed understanding of the CPAR model, please refer to Section \ref{sec_CPAR_basic}). Although the CPAR model can handle both regular and irregular multi-sequence data, its synthetic output for credit card transaction data still leaves room for enhancement in terms of real data fidelity and machine learning utility. Our experiments shed light on the criticality of "pre-processing" in tabular data fields, which markedly influences the fidelity and utility of the synthesized credit card transactions. Such discoveries emphasize the importance of thorough data preparation to elevate the efficacy of generative models, especially when handling intricate datasets like credit card transaction records.

We adopt a \textit{data-centric} strategy to bolster the fidelity and utility of synthetic credit card transaction data. Instead of tweaking the generative model to enhance the quality of synthetic data, we place our primary emphasis on the \textit{preprocessing of raw tabular data}. This shift in focus aims to amplify both the fidelity and utility of the derived synthetic time series. We propose five unique preprocessing schemas, each designed to iteratively enhance the fidelity of the generated synthetic credit card data. After attaining a satisfactory fidelity benchmark, we pivot our efforts toward training fraud detection models \cite{cheng2022downstream, Roy2018DeepLD, Modi2017ReviewOF} specifically tailored for time-series data. Three models — \texttt{XGBoost} \cite{Chen2016XGBoostAS}, \texttt{LGBM} \cite{Ke2017LightGBMAH}, and \texttt{Catboost} \cite{Dorogush2018CatBoostGB} — were chosen and trained on data stemming from the most promising preprocessing schema. Our exploration centers around the efficacy of these models, with a special focus on their False Positive Rate (FPR) and False Negative Rate (FNR) in the fraud detection context (for a detailed analysis, please see Section \ref{sec:result_utility_fraud_detection}). Through this comprehensive approach, we accentuate not just the pivotal role of preprocessing in the realm of synthetic time series data generation, but also its broader ramifications for ensuing machine learning tasks, laying a solid groundwork for ensuing investigations in this area.

\textbf{Paper Organization.} This paper is structured as follows.
Section \ref{sec:relatedWork} gives a review on related keywords about synthetic credit card transaction data. 
Section \ref{sec:method} give comprehensive details on how to preprocess the credit card dataset to train CPAR model for high-fidelity and high-utility. Section \ref{sec:result} gives results on the fidelity evaluation of categorical variables (Section \ref{sec:result_fidelity_categorical}) and continuous variable (Section \ref{sec:result_fidelity_continuous}) and also utility evaluation of fraud detection model performance (Section \ref{sec:result_utility_fraud_detection}). Section \ref{sec:conclusion} talks about our conclusion, new concerns and future direction.

\section{Relate Work}
\label{sec:relatedWork}

\subsection{Synthetic Transactions Dataset}

In recent years, the generation of synthetic data has emerged as a popular research direction \cite{Altman2019SynthesizingCC, Ramachandran2023FraudAmmoLS, VegaMrquez2019CreationOS}, primarily due to the increasing accessibility and ease of use of various tabular synthesis models. These models encompass methodologies based on Generative Adversarial Networks (GANs) \cite{Xu2018SynthesizingTD, Zhao2021CTABGANET, Zhao2022CTABGANET, Lee2022InvertibleTG, Zhao2022FCTGANET}, Diffusion Models \cite{Kotelnikov2022TabDDPMMT, Kim2022STaSyST}, as well as Transformers \cite{ Gorishniy2021RevisitingDL, Borisov2022LanguageMA, Solatorio2023REaLTabFormerGR}, all of which have achieved success across diverse domains of tabular data. Nevertheless, the synthesis of transactional data poses unique research challenges. Transactional data inherently constitutes a time series, and the successful applications of synthetic data models have predominantly been on tabular data lacking temporal stamps. This discrepancy highlights the complexity and specificity required in handling time-series transactional data, necessitating further investigation and innovation in this domain.

\subsection{CPAR model}

The \textit{Conditional Probabilistic Auto-Regressive}, or the CPAR model \cite{Zhang2022SequentialMI} is the main model that aims to capture sequential dependencies currently provided by the \textit{Synthetic Data Vault} (SDV) \cite{Patki2016TheSD}. The model, which is based on neural network, achieves this by capturing inter-row dependencies conditioned on the previous sequence history, and then outputting the necessary parameters to synthesize the future entries. For training, this model also has 3 different loss functions, with each applied to 3 different types of data: continuous numerical, discrete numerical, and categorical.

\subsection{Credit Card Transaction Dataset}

We use credit card transaction dataset provided in \cite{Altman2019SynthesizingCC}. According to the offical documentation of \textit{Synthetic Data Vault} (SDV), the CPAR model is suitable for multi-sequence data, which fits our credit card transaction dataset where there exists multiple users, each with their own transaction history. In addition, we were curious about CPAR's ability to generate high fidelity synthetic data given columns that have continuous numerical data, as well as columns that have high cardinality. Furthermore, the existence of the \texttt{Is Fraud?} column gave us the ability to generate synthetic data to train machine learning models for downstream tasks such as fraud detection. Overall, we believe that this dataset is great for both assessing the quality of the CPAR model, as well as the machine learning efficacy of synthetic dataset for downstream tasks.

\section{Approach and Evaluation Framework}
\label{sec:method}

This section gives comprehensive details on the credit card transaction dataset (Sec. \ref{sec_dataset_basic}), training CPAR model (Sec. \ref{sec_CPAR_basic}), how to preprocess the credit card transaction dataset (Sec. \ref{sec_Preprocessing}) and how to train the fraud detection model with synthetic data (Sec. \ref{sec_Train_FDM}).

\begin{figure}[t]
    \centering
    \includegraphics[width=0.99\textwidth]{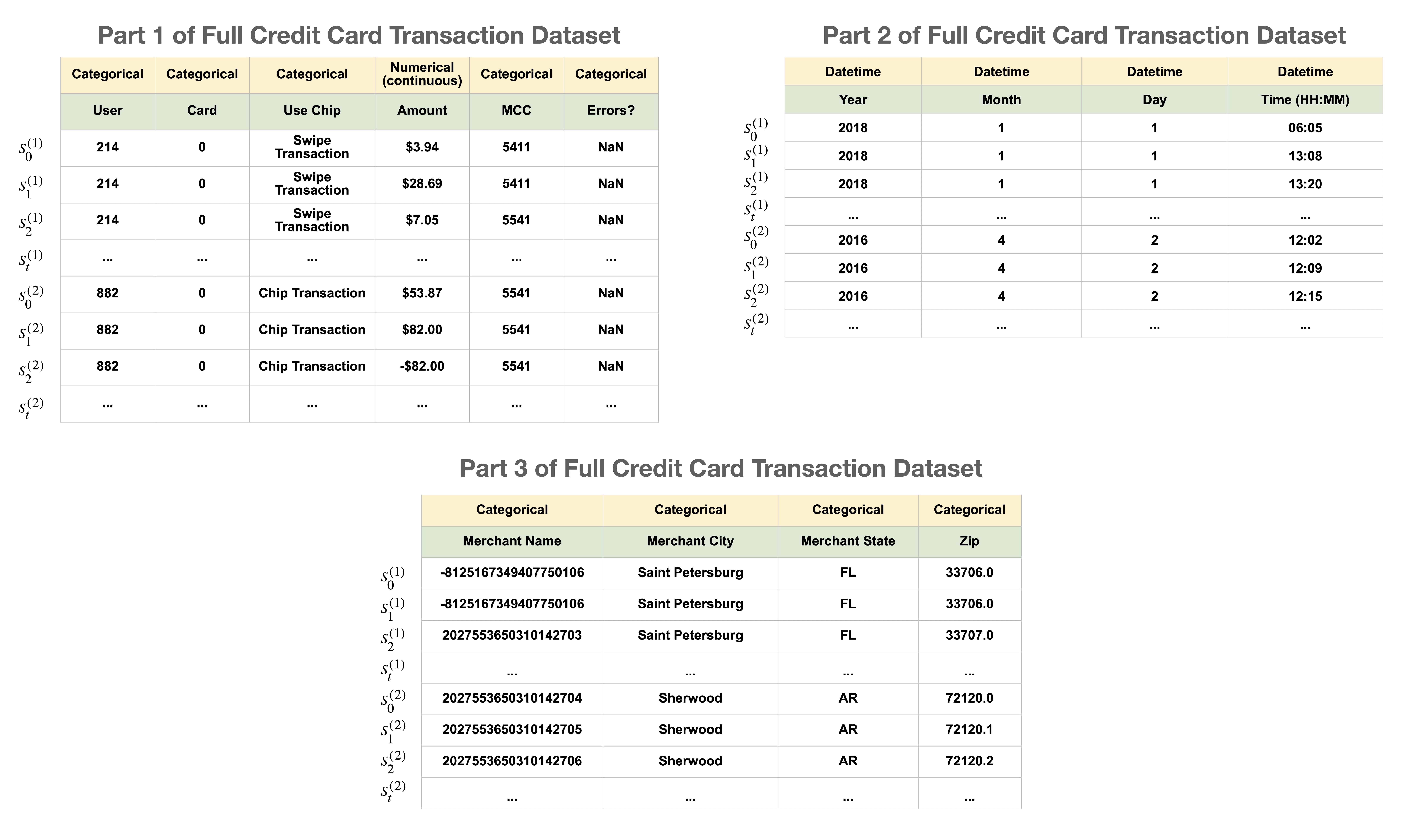}
    \caption{Metadata Data Type of Original Credit Card Transaction Dataset. $S_{j}^{(i)}$ denotes the ith user's jth row.}
    \label{fig:enter-label}
\end{figure}

\subsection{Basic of Credit Card Transaction Dataset}\label{sec_dataset_basic}

Figure \ref{fig:enter-label} gives the full table of the Credit Card Transaction Dataset. The figure, containing 3 parts, shows an overview of the original dataset and its columns, as well as what types of data they are considered under the CPAR framework. In part 1, the figure shows that for columns \texttt{User}, \texttt{Card}, \texttt{Use Chip}, \texttt{MCC}, and \texttt{Errors?}, the CPAR framework considers them as categorical data columns. The \texttt{Amount} column, on the other hand, is considered as a continuous numerical data column. 

In part 2, the figure shows that the columns \texttt{Year}, \texttt{Month}, \texttt{Day}, and \texttt{Time} are not considered as any of the 3 data types (continuous numerical, discrete numerical, or categorical). Rather, they are considered as 'datetime', which are used to set the sequence index under the metadata of the CPAR model. 

In part 3, the figure shows that the columns \texttt{Merchant Name}, \texttt{Merchant City}, \texttt{Merchant State}, and \texttt{Zip} are considered as categorical data columns. Note this is still the case when data in \texttt{Merchant Name} are of type 'integer', as the original dataset follows the same form for this column. This also applies for the \texttt{Zip} column, where all zip codes are displayed as 'float'. In the synthesis of the original data (except for schema 1), we transform the data in these 4 columns into 'string' and consider them as categorical columns when training the CPAR model.

\subsection{CPAR Basic}\label{sec_CPAR_basic}

To train the CPAR model on a dataset of credit card transactions, we categorize the loss functions according to the data type of the variables involved: continuous numerical and categorical.

\textbf{Continuous Numerical Data}:
Consider the parameters $\mu = \pi^{(i)}_{(t, \mu)}$, $\sigma = \pi^{(i)}_{(t, \sigma)}$, and $m = \pi^{(i)}_{(t, m)}$, where $i$ represents the sequence number (or user), and $t$ denotes the transaction index for a given user. The loss function $L(x; \mu, \sigma, m)$ is defined as:
$\mathcal{L}(x ; \mu, \sigma, m)=-\left(\log \left(f_{\mu, \sigma^2}(x)\right)+\log (1-m)\right): x$ is not missing; $\mathcal{L}(x ; \mu, \sigma, m)=-\log (m): x$ is missing.
Here, $\mu$ and $\sigma^2$ represent the mean and variance of a Gaussian distribution, and $m$ indicates the probability of the value being missing. In our case, the credit card transaction dataset contains a continuous numerical column, \texttt{Amount}, parameterized by:
$\pi^{(i)}_{(t, \mu)},  \pi^{(i)}_{(t, \sigma)},  \pi^{(i)}_{(t, m)}$ where $i \in \{0, 1, 2\}$, $t \in \{2540, 2676, 2630 \}$, and $m \in \{0, 0, 0 \}$ corresponding to users 214, 882, and 1798 respectively.

\textbf{Categorical Data}:For categorical data, the loss function is defined as:
$$L(x; \pi_0, \pi_1, \dots, \pi_{N - 1}) = -\sum_{j \in N} x_j \log(\pi_j)$$
where $N$ is the number of categories, $\pi_j$ represents the proportion of category $j$ in the dataset, and $x_j$ is a binary indicator, equal to 1 if the instance belongs to category $j$ and 0 otherwise. In our dataset, the following columns are treated as categorical:
\texttt{Card}, \texttt{Use Chip}, \texttt{Merchant Name}, \texttt{Merchant City}, \texttt{Merchant State}, \texttt{Zip}, \texttt{MCC}, \texttt{Errors?}, and \texttt{Is Fraud?}. Their parameters are denoted as $\pi^{(i)}_{(t, j)}$, with the index sets being: $i \in \{0, 1, 2\}, t \in \{2540, 2676, 2630\}, j = 0$ and the category counts, $N$, detailed as follows:
\begin{enumerate}
    \item \texttt{Card}: $N = 1$,
    \item \texttt{Use Chip}: $N \in \{2, 3, 3 \}$,
    \item \texttt{Merchant Name}: $N \in \{185, 151, 231\}$,
    \item \texttt{Merchant City}: $N \in \{68, 78, 115 \}$,
    \item \texttt{Merchant State}: $N \in \{20, 23, 29 \}$ (excluding NaN),
    \item \texttt{Zip}: $N \in \{87, 86, 134 \}$ (excluding NaN),
    \item \texttt{MCC}: $N \in \{63, 62, 73 \}$,
    \item \texttt{Errors?}: $N \in \{5, 4, 2 \}$ (excluding NaN),
    \item \texttt{Is Fraud?}: $N = 2$.
\end{enumerate}

\textbf{Overall Loss Function.}
With the individual loss functions specified, the overall loss for the model is calculated as: 
\vspace{-2mm}
\begin{equation}\label{eqn:cpar loss}
    \mathcal{L} = \sum_i \sum_t \sum_{j = 0}^{k - 1} \mathcal{L} \left(S^{(i)}_t, \pi^{(i)}_{t, j} \right)
\end{equation}
where $k$ represents the total number of variables.

\textbf{Training Epoch.}
A training epoch for our model, outlined in Algorithm \ref{CPAR train alg}, involves iterating over the sequences and time steps, updating the model parameters, and calculating the loss at each step.
\begin{algorithm}[t]
\caption{One training epoch}\label{CPAR train alg}
    \begin{algorithmic}
        \State $Loss: \mathcal{L}, \enspace Neural \enspace Network: NN$
        \For{each sequence $S^{(i)}$}
            \State $C^{(i)} \gets \textbf{Context}(S^{(i)})$
            \For{each step $S^{(i)}_t$ in $S^{(i)}$}
                \State $\boldsymbol{\pi}^{(i)}_t \gets \textbf{NN}(C^{(i)}, S^{(i)}_0, \ldots, S^{(i)}_{t-1})$
                \State $L \mathrel{+}= L(S^{(i)}_t; \boldsymbol{\pi}^{(i)}_t)$
            \EndFor
        \EndFor
        \State $NN \gets \min(L, NN)$
    \end{algorithmic}
\end{algorithm}


For Algorithm \ref{CPAR train alg}, the loss function $\mathcal{L}$ is defined as \eqref{eqn:cpar loss}, and the neural network is built with 4 layers: a GRU layer in between 2 dense layers, alongside with a final layer with different activation functions depending on the column data type. For more details, see section 3.1.2 of \cite{Zhang2022SequentialMI}. 

Next, for each sequence $S^{(i)}$, the constant input \textbf{Context} is assigned to the variable $C^{(i)}$. Then, for each row $S^{(i)}_t$, $C^{(i)}$ and all $S^{(i)}_{0}, \dots, S^{(i)}_{t - 1}$ are inputted into the neural network to output parameters $\pi^{(i)}_{(t, 0)}, \pi^{(i)}_{(t, 1)}, \dots$. The training loss are then calculated from the parameters and added to the total loss. 
Finally, the loss function for the neural network is then minimized and outputted. For more details, see section 3.1 of \cite{Zhang2022SequentialMI}.

\begin{figure}[t]
    \centering
    \includegraphics[width=1\textwidth]{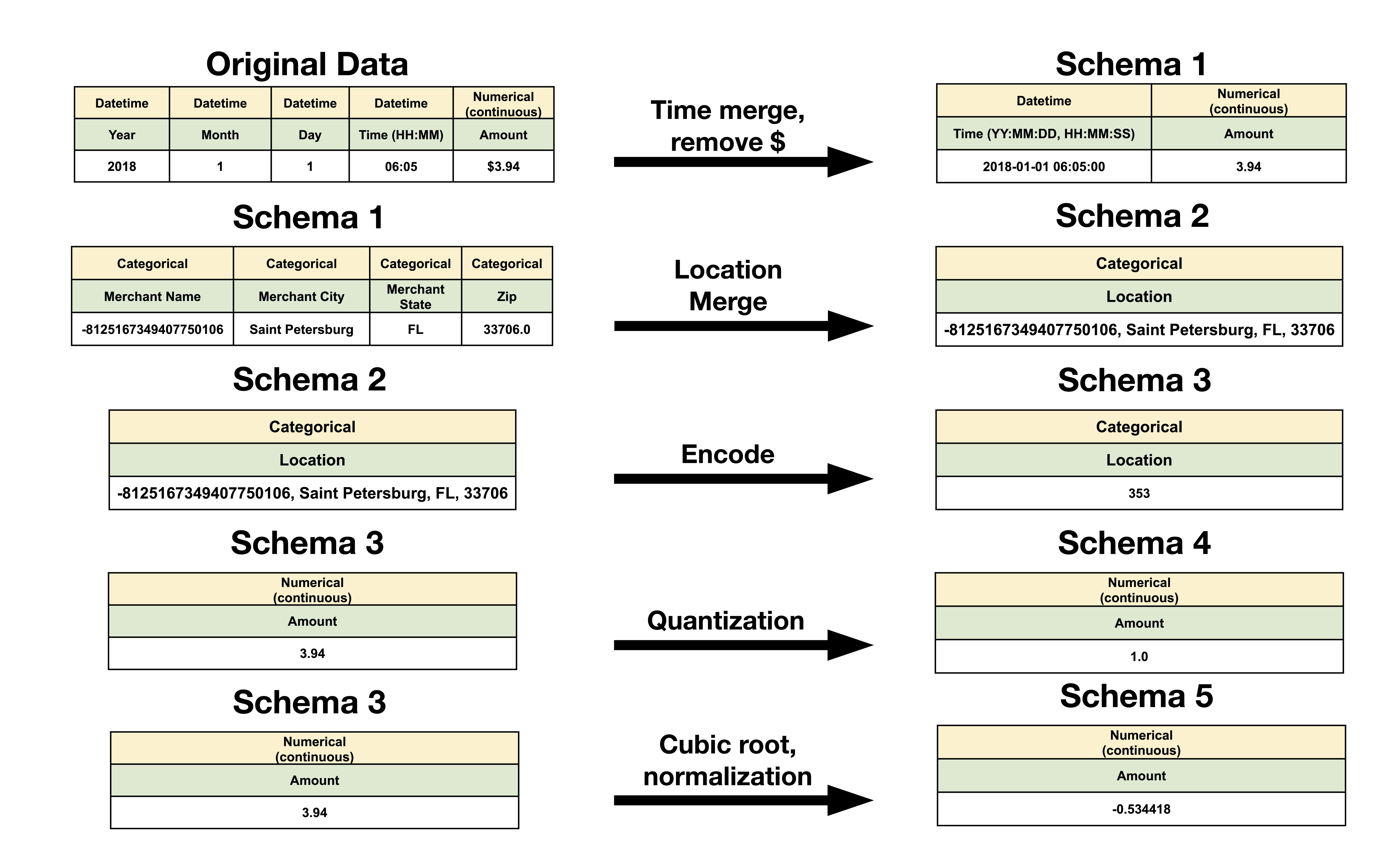}
    \caption{Differences of Columns between Schemas}
    \label{fig:schema diff}
\end{figure}

\subsection{Preprocessing Credit Card Transaction Dataset}\label{sec_Preprocessing}

We present a variety of preprocessing schemas to generate synthetic credit card transaction datasets via the CPAR model, each introducing unique modifications for comprehensive analysis. The initial dataset and metadata configurations serve as the foundation for each schema.


\textbf{Schema 1: Minimal CPAR Requirements.} To meet the minimum requirements for running the CPAR model, we focus on three users and transform the dataset as follows: combine \texttt{Year}, \texttt{Month}, \texttt{Day}, and \texttt{Time} into a single Pandas datetime column; convert \texttt{Amount} to a numerical format, removing the \$ sign; replace NaNs in \texttt{Zip} with 'not applicable' for non-U.S. zip codes, converting valid zips to 'string'; change NaNs in \texttt{Errors?} to 'none', indicating error-free transactions; adjust \texttt{Merchant Name} to string, and \texttt{Is Fraud?} to boolean. With metadata from the preprocessed data, we apply CPAR (parameters: \texttt{epochs} = 1024, \texttt{context\_columns} = []) to generate the initial synthetic dataset.


\textbf{Schema 2: Handling Location and Avoiding Non-Existent Entries.} Building on the previous schema, we integrate \texttt{Merchant Name}, \texttt{Merchant City}, \texttt{Merchant State}, and \texttt{Zip} into a single 'string' data type \texttt{Location} column to \textit{prevent the creation of non-existent locations}, a common issue in Schema 1 due to possible shifts in these four columns. With updated metadata marking \texttt{Location} as 'categorical', we proceed to generate the second synthetic dataset using the same CPAR parameters.


\textbf{Schema 3: Enhancing Machine Learning Efficacy through Categorical Encoding.} Building upon prior schemas, we apply Scikit-Learn’s Label Encoder to categorical columns (\texttt{Use Chip}, \texttt{MCC}, \texttt{Errors?}, \texttt{Location}, and \texttt{Is Fraud?}), maintaining their 'string' data types and unchanged metadata. This strategy ensures CPAR’s accurate recognition of these as categorical, avoiding the potential generation of out-of-category values in the case of 'integer' data types. Post-encoding, we generate the third synthetic dataset and revert the columns to their original states using the trained encoders.

\textbf{Schema 4: Addressing Non-Gaussianity with Quantile Transformation.} Extending previous steps, Schema 4 employs quantile encoding for the \texttt{Amount} column, transforming its values logarithmically into 10 bins, following \cite{Padhi2020TabularTF}. Post-transformation, this column is converted to 'string' to generate the fourth synthetic dataset, before undergoing an inverse transformation to revert changes.

\textbf{Schema 5: Mitigating Non-Gaussianity with Cubic Root Transformation.} Building on Schema 3, this final schema standardizes the \texttt{Amount} column to zero mean and unit variance, followed by a cubic root transformation. This prepares the data for the fifth synthetic dataset generation, after which all transformations and scalings are reversed to preserve data integrity.

Each schema enhances the dataset's structure and content, facilitating a thorough analysis through the CPAR model.

\subsection{Training Fraud Detection Model}\label{sec_Train_FDM}

This subsection outlines the process of training a fraud detection model, detailing each step from data preparation to performance evaluation.

\begin{itemize}
    \item \textbf{Step 1: Synthetic Data Generation.} Utilizing schema 5, we generate synthetic datasets for 3 users with lengths determined by the CPAR model, choosing the first dataset for our machine learning efficacy experiment.
    \item \textbf{Step 2: Data Categorization.} We partition the datasets into fraud and non-fraud samples, subsequently using CPAR to create a synthetic dataset comprising 15,000 fraud transactions.
    \item \textbf{Step 3: Dataset Preparation.} Five datasets, each with 10,000 samples and varying fraud-to-non-fraud ratios (1\%, 5\%, 10\%, 20\%, and 50\%), are prepared. The \texttt{Time} column is dropped, and a \texttt{time\_diff} column is added to indicate \textit{the time since the last event in minutes}, initialized to 0 for the first transaction of each user.
    \item \textbf{Step 4: Model Training.} We train models using the prepared datasets, employing metadata from \ref{sec_dataset_basic}, $context = []$, and setting epochs to 1024. 
    \item \textbf{Step 5: Performance Evaluation.} Model performance is assessed by comparing accuracy (False Positive Rate and False Negative Rate) across the five datasets using original data as ground truth.
\end{itemize}

Through these steps, we conduct a thorough evaluation of our fraud detection models, ensuring that they are well-suited to handle the complexities and challenges of identifying fraudulent activities in transactional data. The use of synthetic data, coupled with careful data preparation and categorization, ensures that our models are trained on relevant and representative data, providing a solid foundation for accurate and reliable fraud detection.

\textbf{Training on original and encoded data.} For measuring machine learning efficacy, not only did we trained with untransformed data, but also did we consider encoded data. We label encoded every categorical column in the synthetic datasets and added identical \texttt{{time\_diff}} column to the datasets. Then, we proceeded to train new machine learning models with these synethetic datasets, and compared the performance of machine learning models trained on encoded and un-encoded data.

\textbf{Fraud Detection Model Class.} To conduct a comprehensive efficacy analysis, we have implemented three distinct fraud detection models, each designed to handle varying fraud-to-non-fraud ratios. These models are selected for their unique strengths and capabilities in tackling different aspects of fraud detection, ensuring a robust and versatile evaluation.

\textbf{(I) Categorical Boosting (Catboost)}~\cite{Dorogush2018CatBoostGB} CatBoost is a powerful algorithm renowned for its proficiency in managing categorical features directly without necessitating preprocessing, which can be a game changer in fraud detection where data is often diverse and complex. In our setup, we specify the 'categorical\_feature\_indices' parameter to ensure that the model accurately recognizes and utilizes the categorical data within the dataset. We follow a standard 80\%-20\% train-test split to assess the model’s performance and prevent overfitting. In terms of hyperparameters, we have settled on 500 iterations, a learning rate of 0.01, and a tree depth of 5, ensuring a balanced trade-off between model complexity and training efficiency. The 'Logloss' loss function is employed to optimize the model’s performance, and we set a random seed of 200 to guarantee reproducibility in our results. The model’s effectiveness is meticulously validated using the original dataset to ensure authenticity and reliability in its fraud detection capabilities.
 
 \textbf{(II) Extreme Gradient Boosting (XGBoost)}~\cite{Chen2016XGBoostAS} XGBoost stands out for its scalability and computational efficiency, which is paramount in fraud detection due to the high volume of transactions that need to be analyzed promptly. We harness the power of the 'hist' tree method in XGBoost, which is known for its faster computation times and efficient memory usage, coupled with its ability to handle categorical inputs seamlessly. After splitting our data following the 80\%-20\% rule for training and testing, we configure the model with hyperparameters akin to those used in CatBoost, albeit with 'binary\:logistic' specified as the objective function to tailor the model for binary classification tasks, which is a common scenario in fraud detection. A different random seed is set to ensure diversity and robustness in our results. As with the previous model, we use the original data to validate the model’s performance, ensuring a thorough and accurate evaluation.
 
 \textbf{(III)Light Gradient Boosting Machine (LGBM)} \cite{Ke2017LightGBMAH} 
    LGBM is renowned for its efficiency and speed, making it an ideal candidate for fraud detection tasks that demand quick and accurate results. It utilizes Gradient-Based Decision Trees (GBDT), Gradient-based One-Side Sampling (GOSS), and Exclusive Feature Bundling (EFB) to enhance its performance and efficiency. Following the standard practice, we split our dataset into 80\% training and 20\% testing portions. The model is then configured with 500 iterations, a learning rate of 0.01, and a 'binary' objective to align with the nature of fraud detection tasks. The 'binary\_logloss' metric is used to assess the model’s performance, ensuring a precise evaluation. A random state of 100 is set for reproducibility. Similar to the other models, we validate LGBM’s performance using the original dataset, ensuring a reliable and accurate assessment of its fraud detection capabilities.

By leveraging the strengths of CatBoost, XGBoost, and LGBM, we aim to provide a comprehensive analysis that addresses various challenges in fraud detection, ultimately contributing to more secure and trustworthy financial transactions. Each model’s unique features are meticulously harnessed to optimize their performance in fraud detection, providing us with valuable insights and a robust evaluation of their effectiveness.

\section{Evaluation Results}
\label{sec:result}

\begin{figure*}[t]
\centering
    \begin{minipage}[b]{0.48\textwidth}
        \centering
        \includegraphics[height=7.25cm]{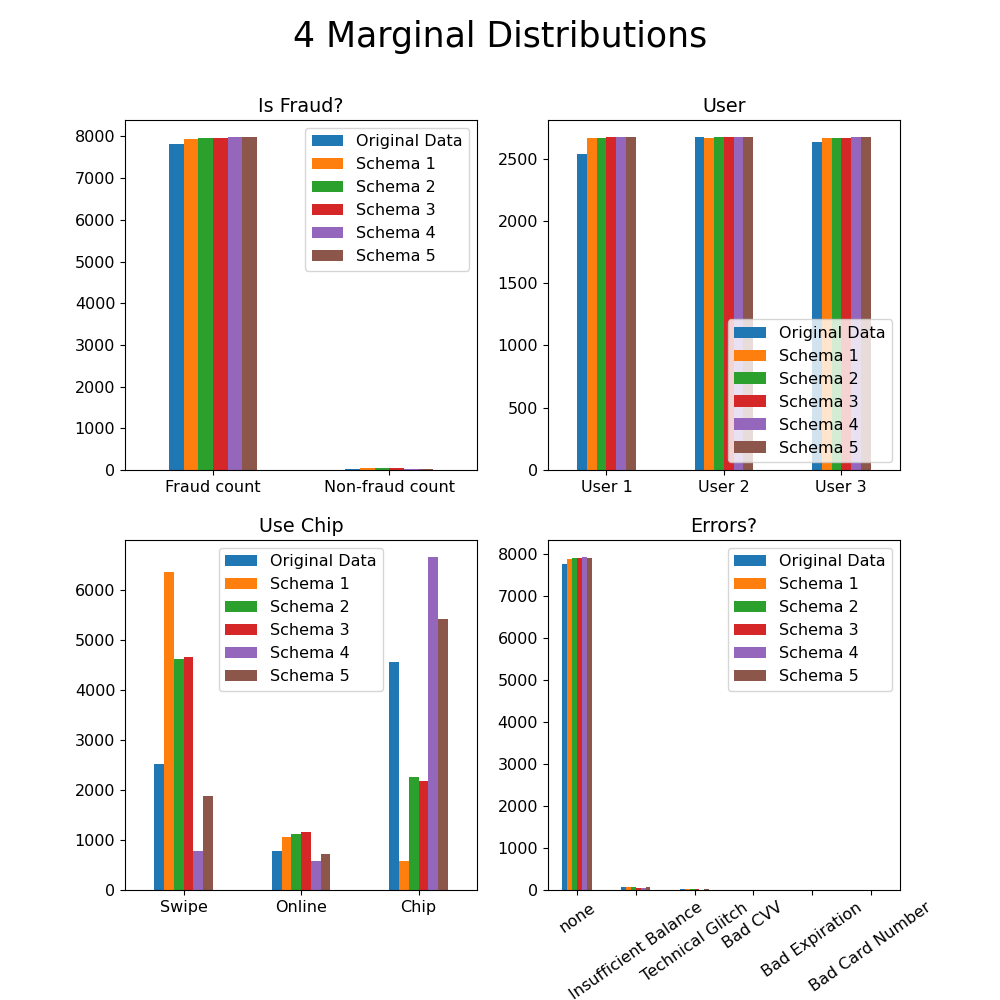}
        \caption{Marginal Distributions of columns \texttt{Is Fraud?, User, Use Chip, Errors?}, from original data set and Schema 1 to 5}
        \label{fig:4_marginal_distributions}
    \end{minipage}\hfill
    \begin{minipage}[b]{0.48\textwidth}
        \centering
        \includegraphics[height=3.5cm, width=8.5cm]{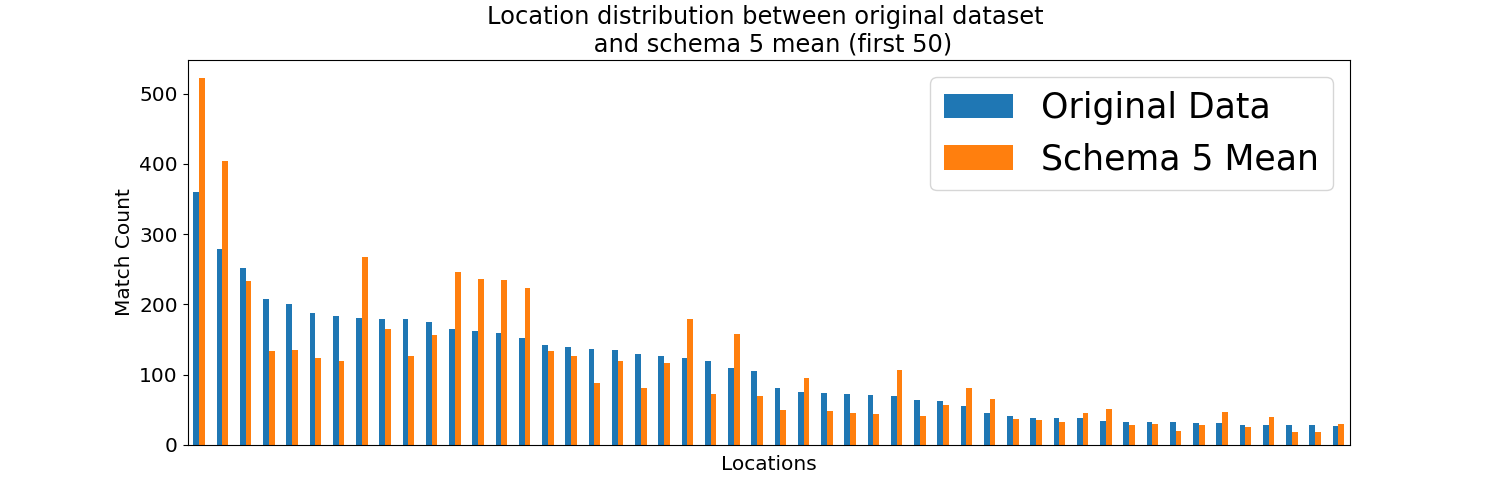}
        \includegraphics[width=0.8\textwidth]{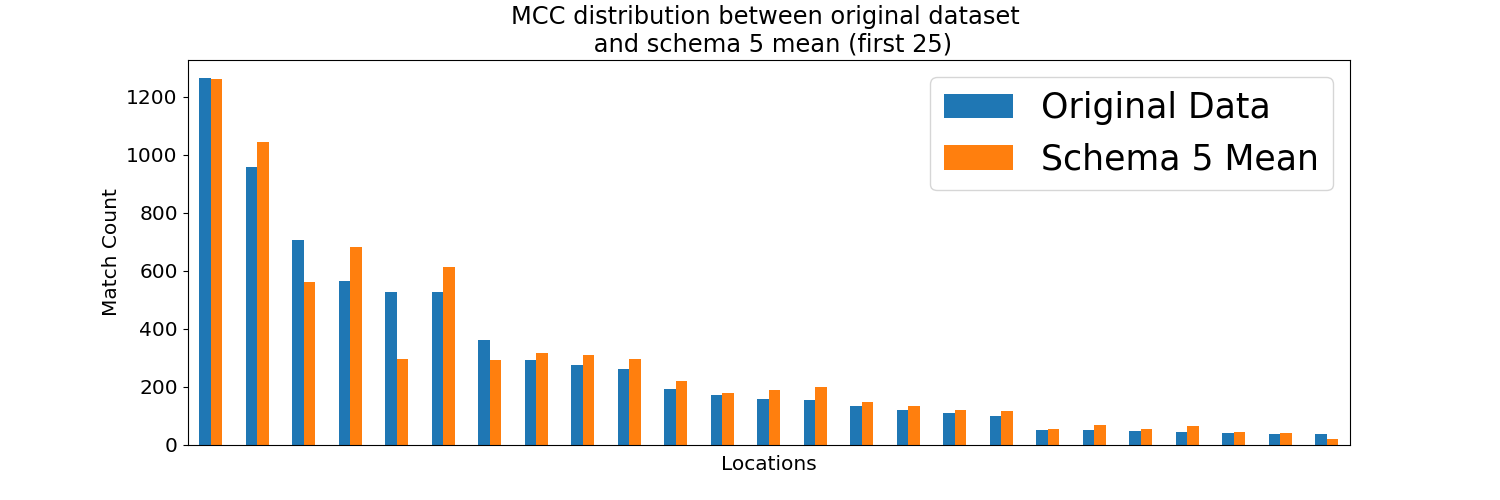}
        \caption{ Marginal Distribution of Location (first 50 entries) for Synthetic dataset, Schema 5}
        \label{fig:fidelity_location}
    \end{minipage}
\end{figure*}

In this section, we give empirical evaluation on the fidelity and utility of resulting synthetic credit car transaction time series. At section \ref{sec:result_fidelity_categorical}, we present fidelity evaluation on categorical variable (\texttt{MCC} \texttt{Errors?}, \texttt{Users}, \texttt{Use Chip}, \texttt{Location}, \texttt{Is Fraud?}). At section \ref{sec:result_fidelity_continuous}, we present fidelity evaluation on continuous variable (\texttt{Amount}). At section \ref{sec:result_utility_fraud_detection}, we present utility evaluation on the fraud detection model trained on the synthetic credit card transaction dataset.

\subsection{Fidelity of Synthetic Categorical Variable}\label{sec:result_fidelity_categorical}

Figure \ref{fig:4_marginal_distributions}, first plot: This plot shows 4 marginal distributions of low-cardinal categorical columns (\texttt{Is Fraud?}, \texttt{User}, \texttt{Use Chip}, \texttt{Errors?}). Specifically, the plots display the counts of each category across the original dataset and all synthetic dataset generated from each schema. We also note that the column \texttt{Card} is not included, as it remains a constant for all datasets. The first plot in figure 3 indicates that for columns \texttt{Is Fraud?}, \texttt{User}, and \texttt{Errors?}, the differences in marginal distributions between the ground truth and all 5 datasets generated from each schema are minimal. For column \texttt{Use Chip}, however, the plot indicates that datasets generated from schema 1, 2, 3, and 4 (colors orange, green, red, and purple, respectively) are visibly farther from the original dataset (blue) than the dataset generated from schema 5 (brown). Thus, \textbf{most synthetic datasets closely match the original, except for "Use Chip".} 

Figure \ref{fig:fidelity_location}, first plot: For the marginal distribution of \texttt{Location}, we first sort the count of unique locations in the original dataset in descending order, then plot out the first 50 comparisons with the mean count across all 20 Schema 5. The second plot in figure 2 indicates that for locations that have higher counts in the original dataset, there exists some discrepancies between the original dataset and the dataset generated from schema 5. Moving to the right of the plot where locations have lower counts, the discrepancies starts to decrease, and the distribution is more similar. Thus, we found \textbf{discrepancies between original and schema 5 datasets decrease for lower-count locations.}

Figure \ref{fig:fidelity_location}, second plot: For the marginal distribution of \texttt{MCC}, we also sort the count of unique MCC in the original dataset in descending order. Then, we plot out the first 25 comparisons with the mean count across all 20 Schema 5. The third plot in figure 2 indicates that the marginal distribution of MCC between the original dataset and the dataset generated from schema 5 are really close to each other. Thus, we found \textbf{The MCC distribution in Schema 5 closely matches the original.}

\begin{figure}[t]
    \centering
    \includegraphics[width=.85\textwidth]{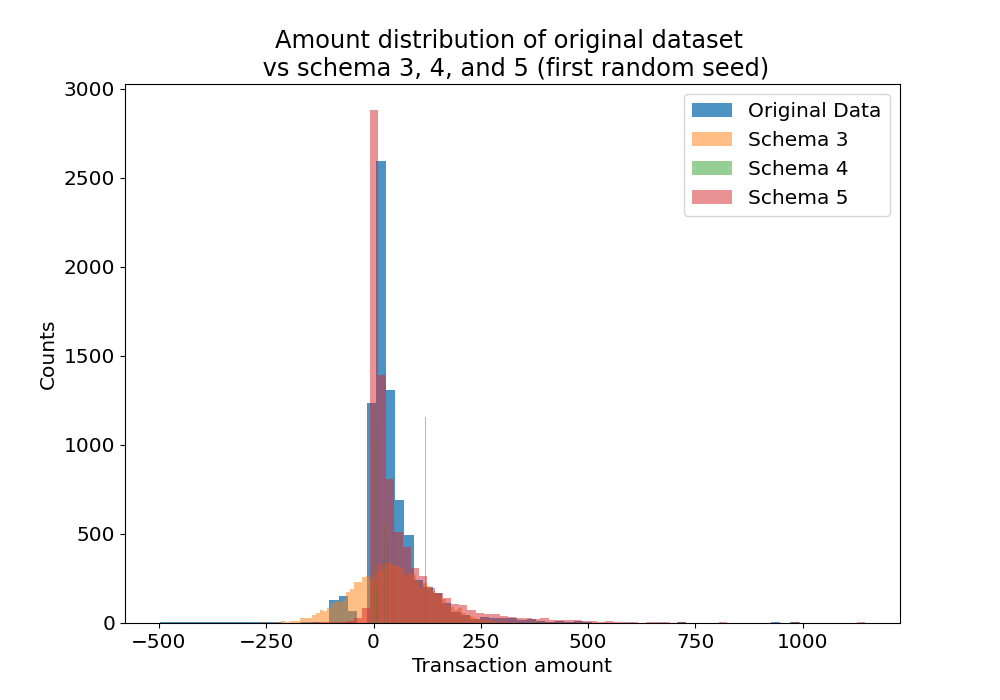}
    \caption{Amount Distribution for Original Dataset vs Synthetic Datasets Generated From Schema 3, 4, and 5}
\label{fig:amount_dist}
\end{figure}

\subsection{Fidelity of Synthetic Continuous Variable}\label{sec:result_fidelity_continuous}

Figure \ref{fig:amount_dist} gives the fidelity evaluation result for the continuous variable in the credit card transaction dataset. The figure indicates that dataset generated from schema 3 (before data transformation, in orange), shows a shifted Gaussian distribution, which is accurate from the description of the CPAR model. The dataset generated from schema 4 (green) shows multiple long and thin bars as a result of transaction amount quantization. The dataset generated from schema 5 (red) shows a distribution that is really close with the marginal distribution of the transaction amounts in the original dataset. The plot indicates that between the 3 datasets generated from schema 3, 4, and 5, schema 5 is the closest to the original data by a large margin. Thus, we found \textbf{Schema 5 closely matches the original transaction amount distribution.}

\begin{figure}[t]
\centering
\includegraphics[width=.49\textwidth]{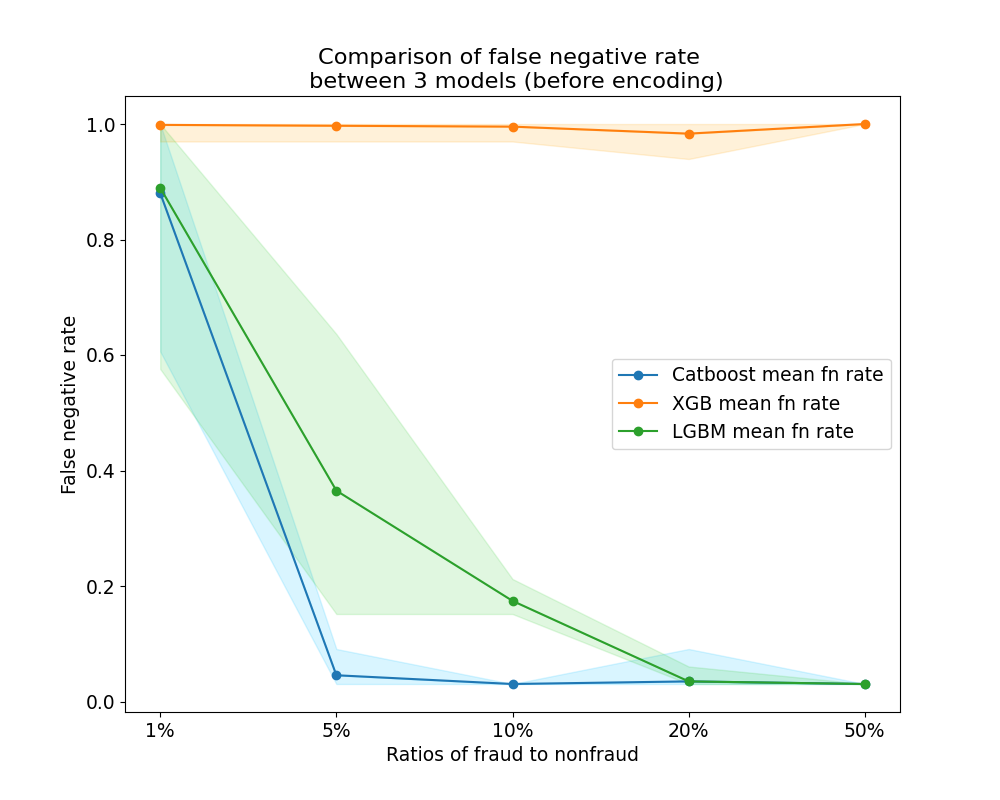}
\includegraphics[width=.49\textwidth]{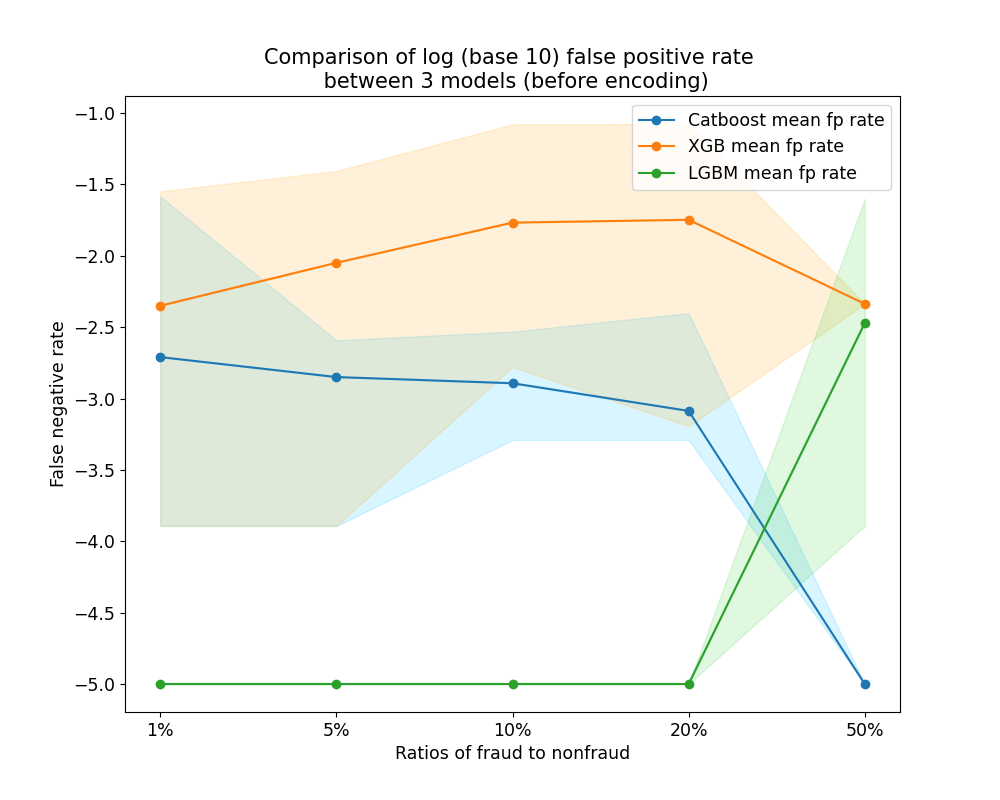}
\caption{False Negative Rate (Left) and False Positive Rate (Right) for Fraud Detection Result of \texttt{Catboost, XGBoost} and \texttt{LGBM}.}
\label{fig:fraud_detection_performance}
\vspace{-3mm}
\end{figure}

\subsection{Utility of Synthetic Data trained Fraud Detection model}\label{sec:result_utility_fraud_detection}

Figure \ref{fig:fraud_detection_performance} gives a utility evaluation of synthetic data trained fraud detection model. Note that since the original dataset is highly imbalanced (0.4206\% fraud transactions and 99.5794\% non-fraud transactions), monitoring the false positive and false negative rate is highly important, as high accuracy does not imply low false positive or false negative rates. 

The left figure indicates that for the false negative rate across all 3 machine learning models, Catboost and LGBM have the best resutls, with Catboost reaching and LGBM reaching 3.0303\% in mean false negative rate across training on 20 different seeds (essentially only missing 1 out of 33 fraud transactions) at 20\% and 50\% fraud to non-fraud ratio. While LGBM suffers from higher uncertainty, we note that training speed of LGBM is way faster than the training speed of Catboost. On the other hand, XGBoost fails to deliver great result, with a mean of almost 100\% across all fraud to non-fraud ratios. This might be caused by the lack of support of categorical features in the library (the support for categorical features is currently still in experimentation). Thus, we found \textbf{\texttt{Catboost} and \texttt{LGBM} outperform \texttt{XGBoost} in false negative rate.}

The right figure indicates that for the log base 10 of the false positive rate across all 3 machine learning models, the false positive rate of Catboost continues to decrease (at last to <0.0001\%) as fraud to non-fraud ratio increases, while LGBM encounters a great leap in false positive ratio when moving from 20\% fraud to non-fraud ratio to 50\% fraud to non-fraud ratio. XGBoost, on the other hand, maintains a high false positive rate throughout all the different fraud to non-fraud ratios. Thus, we found \textbf{\texttt{Catboost} outperform \texttt{LGBM} and \texttt{XGBoost} in false positive rate.}

Lastly, note that the previous models are trained from decoded data, meaning that they are similar to that of the original data. We also trained our models using encoded \textbf{Dataset: Schema 5} and validated on encoded original data, which is the same encoded schemas as \textbf{Dataset: Schema 3 and 5}. The results we discovered is that for mean false negative rate across all fraud to non-fraud ratios, all models remain robust (cite appendix). For mean false positive rate, on the other hand, both XGBoost and Catboost remain robust, while Light-GBM appears to be sensitive to label encoding.Thus, we found \textbf{
\texttt{Light-GBM} is sensitive to label encoding, while \texttt{XGBoost} and \texttt{Catboost} remain robust.}

\section{Conclusion}
\label{sec:conclusion}

We evaluate the fidelity and utility of synthetic data generated from the application of the CPAR model to a real-world Credit Card Transaction dataset. By employing Schema 5 for preprocessing the original dataset, we ensure that the distribution of both categorical and continuous variables is well-preserved in the synthesized time-series data. Based on Schema 5, we generate synthetic data to train three representative fraud detection models (\texttt{Catboost, XGBoost, LGBM}), further investigating the utility of the synthetic data as training material for machine learning. Utilizing the synthetic data allows us to increase the proportion of fraud cases in our training dataset, leading to near-zero False Positive and False Negative Rates for \texttt{Catboost} and \texttt{LGBM}. From the insights gained in this study, we posit that, \textbf{given the appropriate preprocessing schema, synthetic data can indeed serve as a high-fidelity copy of the original data, enhancing the performance of fraud detection by generating additional fraud case data.}

\clearpage

\baselineskip=13pt
\bibliographystyle{plain}
\nocite{*}
\bibliography{ref}

\clearpage
\appendix

\section{Appendix}

This section provides additional empirical results for Implementation of \texttt{CPAR} Model On Credit Card Transaction Data discussed at Section \ref{sec:result}. 

\begin{figure}[ht]
    \centering
    \includegraphics[width=.8\textwidth]{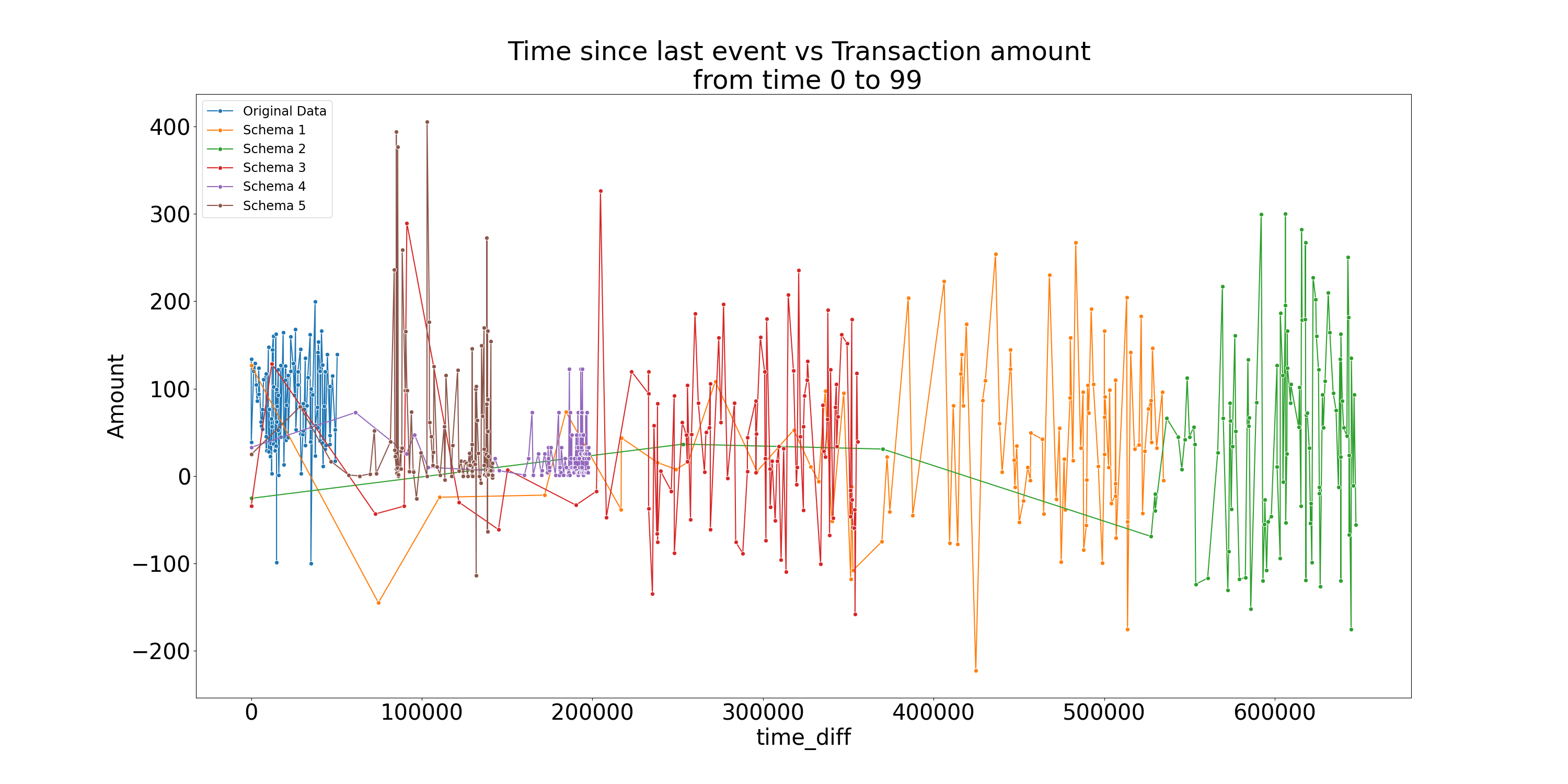}
    \includegraphics[width=.8\textwidth]{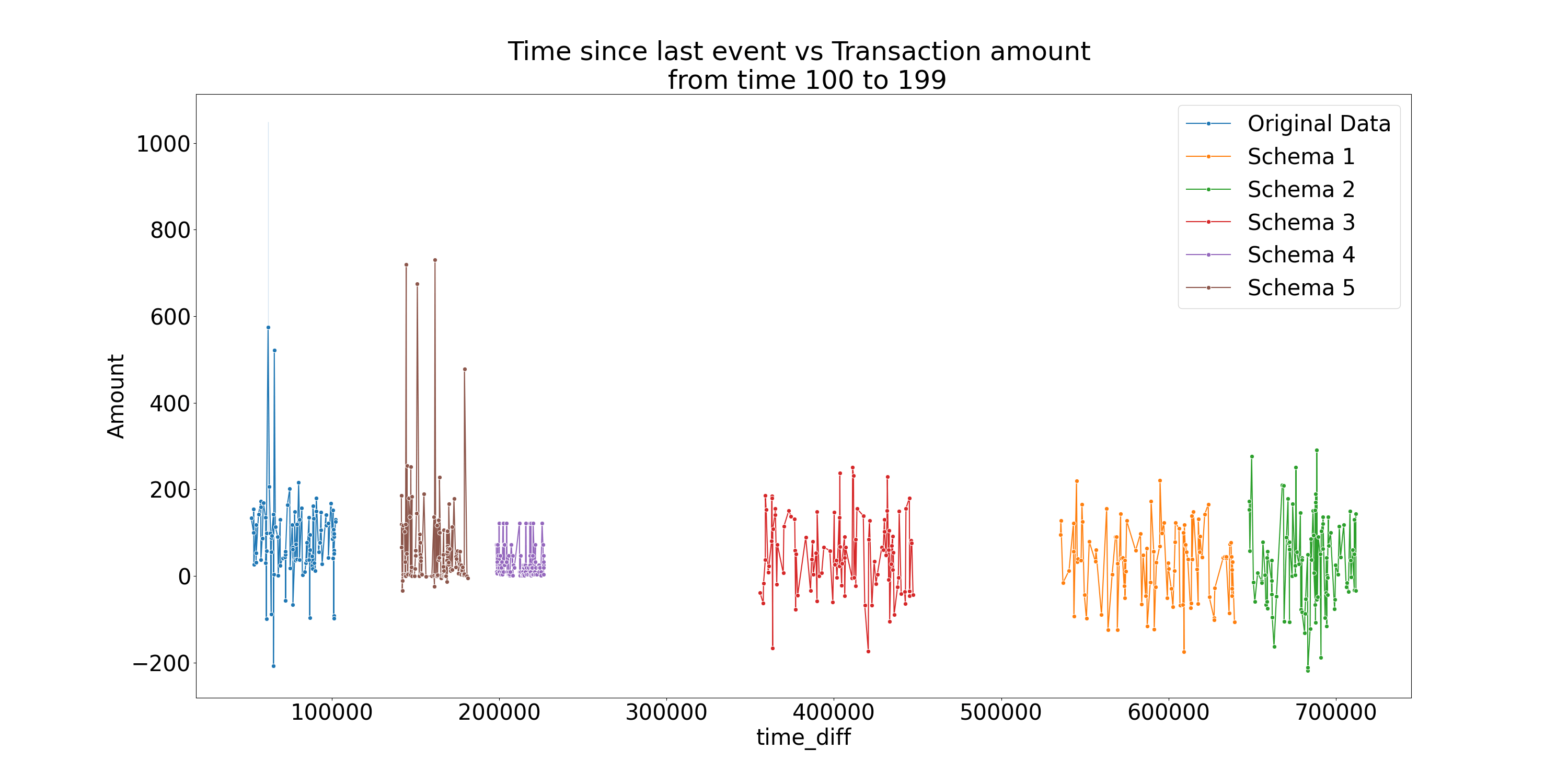}
    \includegraphics[width=.8\textwidth]{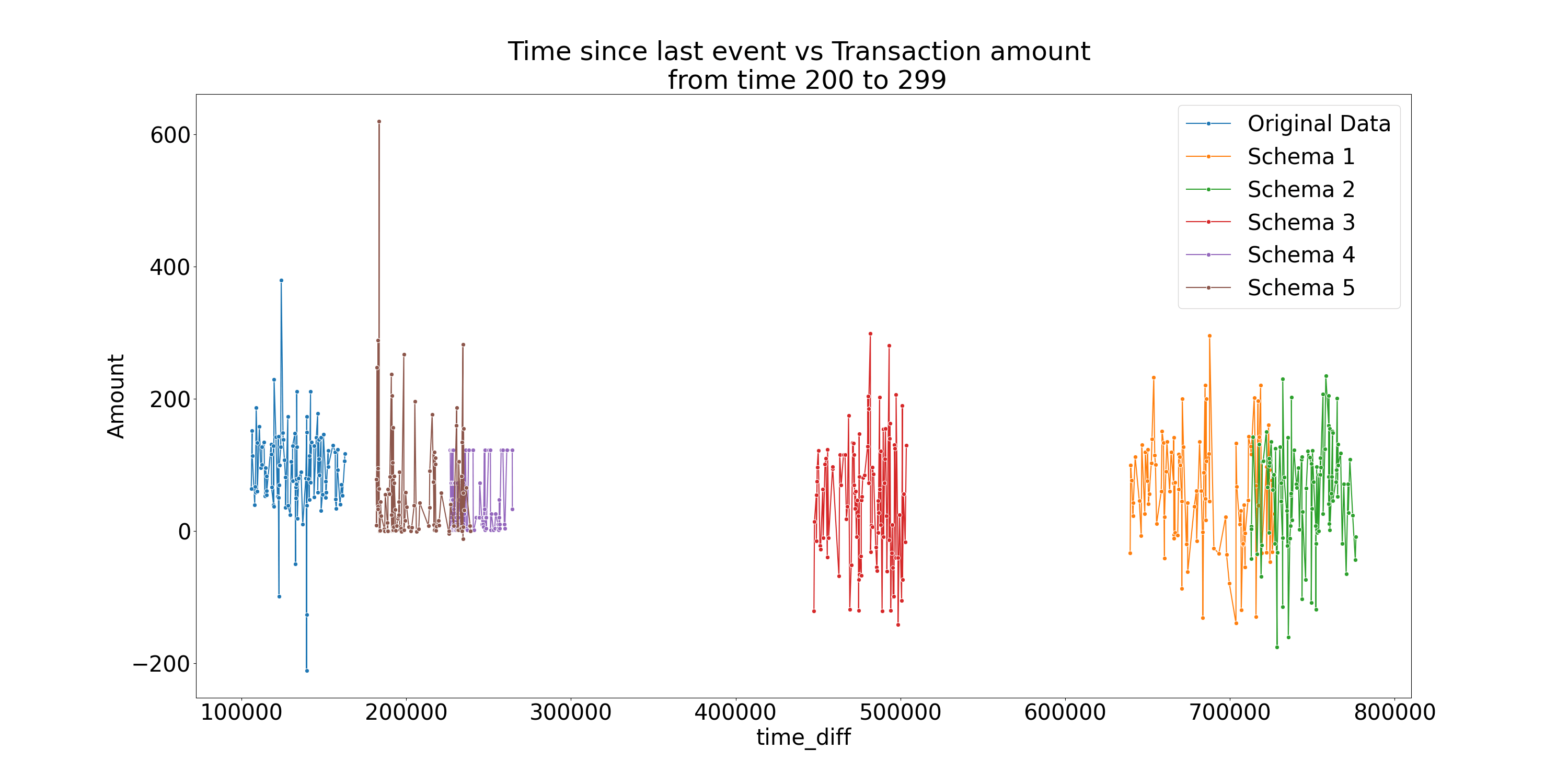}
    \caption{Time Since Last Event vs Transaction Amount (Time Dependency). Even though there are obvious gaps between the time dependency graph of the original data (in blue) to the data from Schema 1 to 5, the distance between the original and Schema 5 (in brown) is the lowest.}
    \label{Appendix:time_dependencies}
\end{figure}

\begin{figure}
\centering
    \begin{minipage}{0.5\textwidth}
        \centering
        \includegraphics[width=1\textwidth]{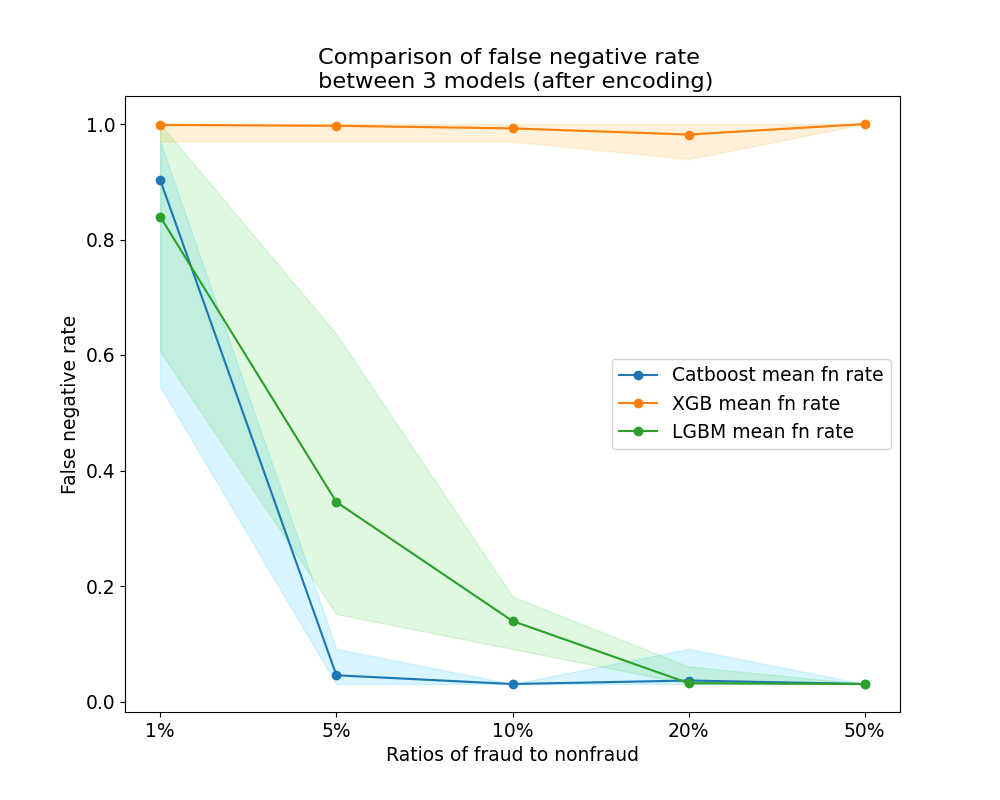} 
        \caption{first figure}
    \end{minipage}\hfill
    \begin{minipage}{0.5\textwidth}
        \centering
        \includegraphics[width=1\textwidth]{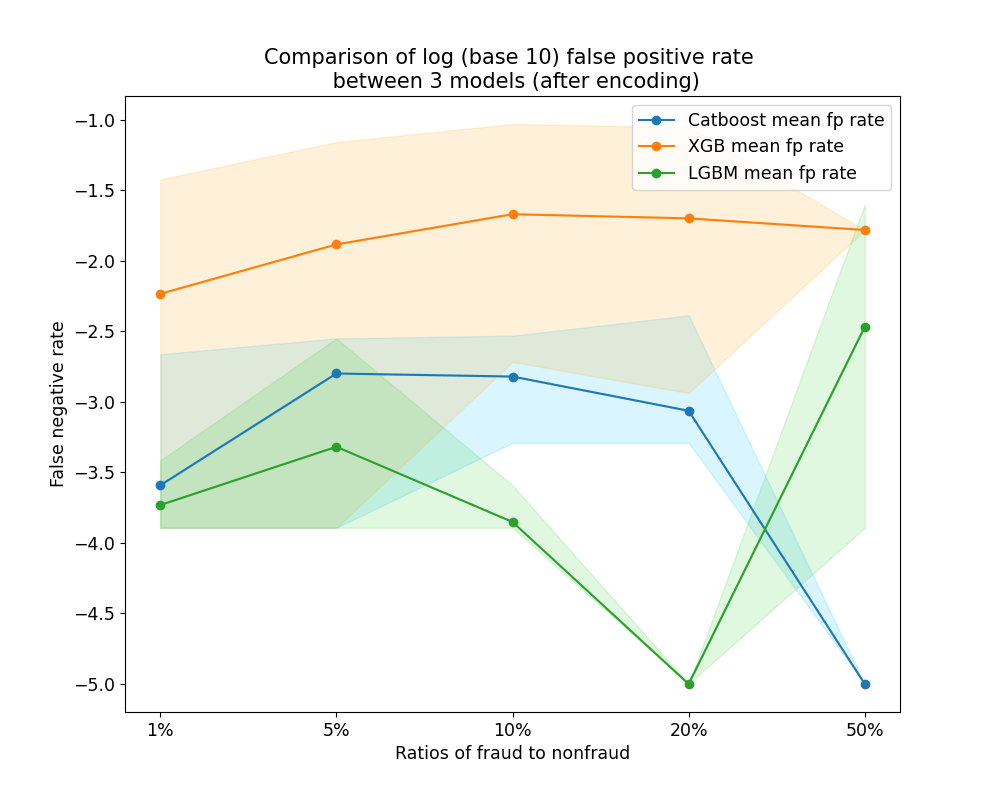} 
        \caption{second figure}
    \end{minipage}
    \caption{FNR and FPR. \texttt{Light-GBM} is found to be sensitive to label encoding, while \texttt{XGBoost} and \texttt{Catboost} remain robust.}
    \label{Appendix:fnr_fpr}
\end{figure}

\begin{figure*}
    \centering
    \includegraphics[width=.9\textwidth]{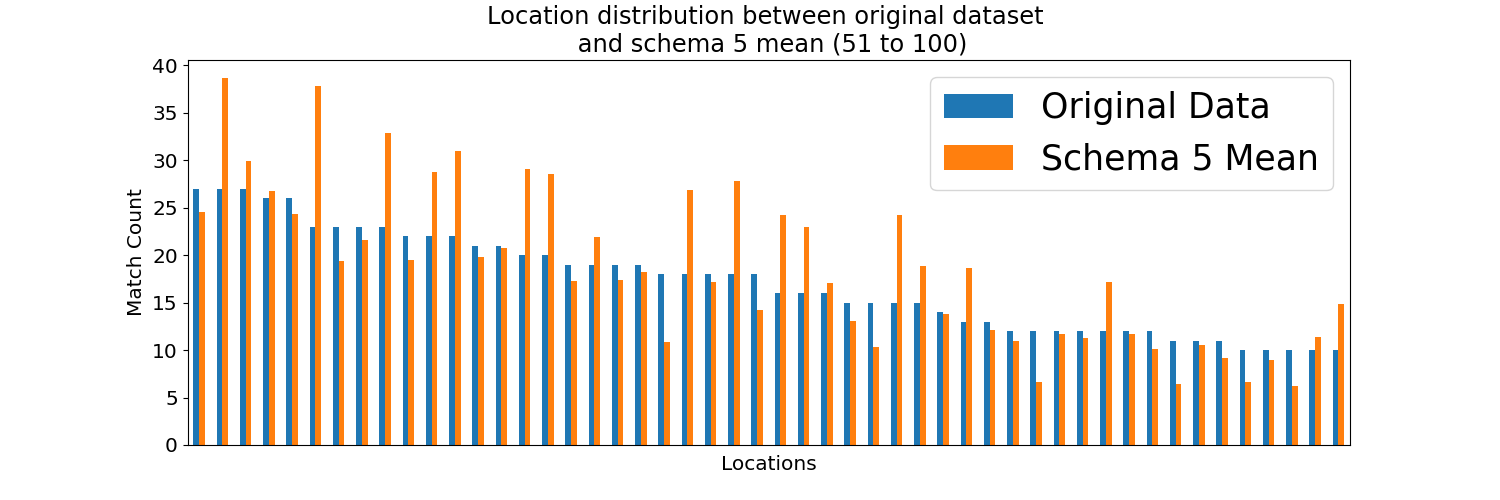}
    \includegraphics[width=.9\textwidth]{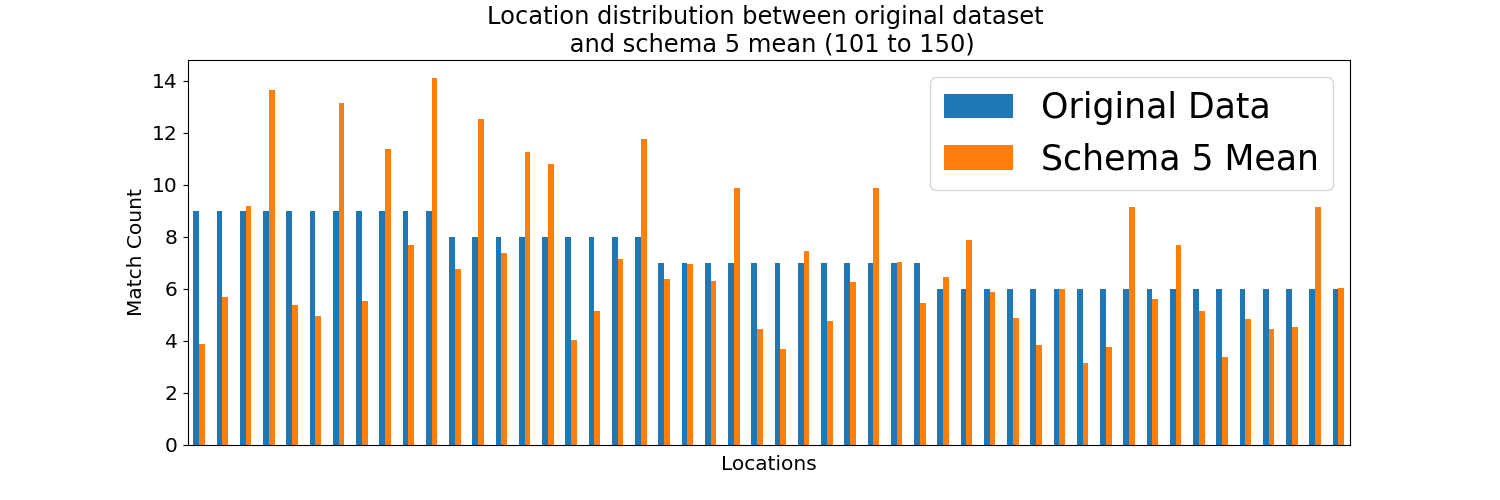}
    \caption{Marginal Distribution of the \texttt{Location} column. The next 100 entries are shown, where after the 150th entry the single digit matching counts become insignificant.}
    \label{Appendix:location_dist}
\end{figure*}

\begin{figure}
    \centering
    \includegraphics[width=.9\textwidth]{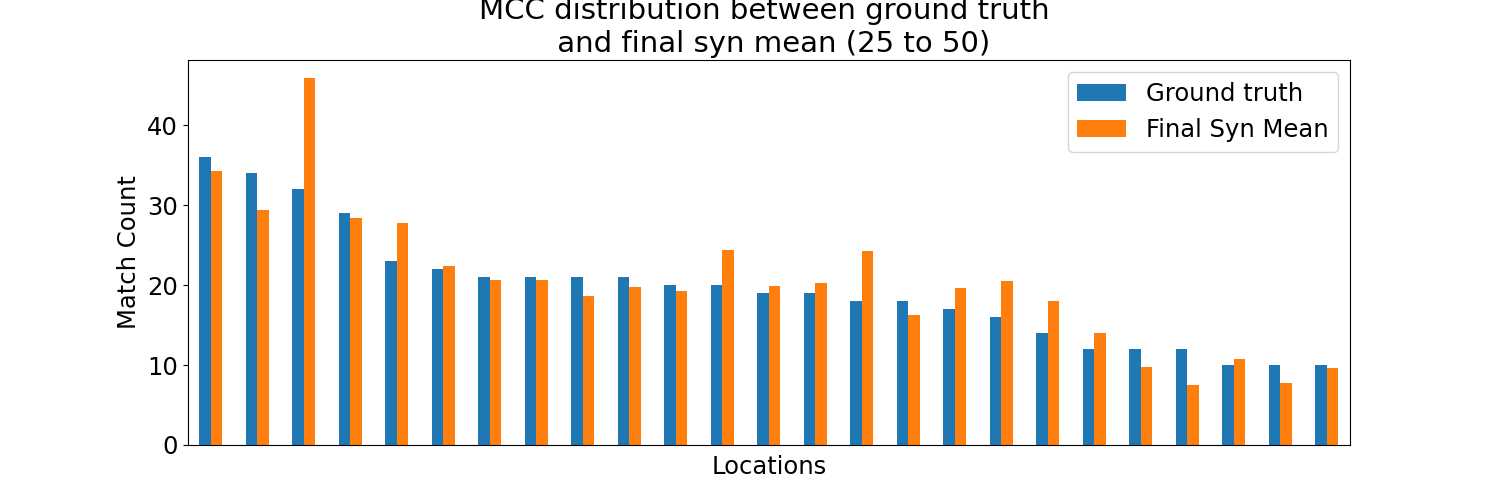}
    \includegraphics[width=.9\textwidth]{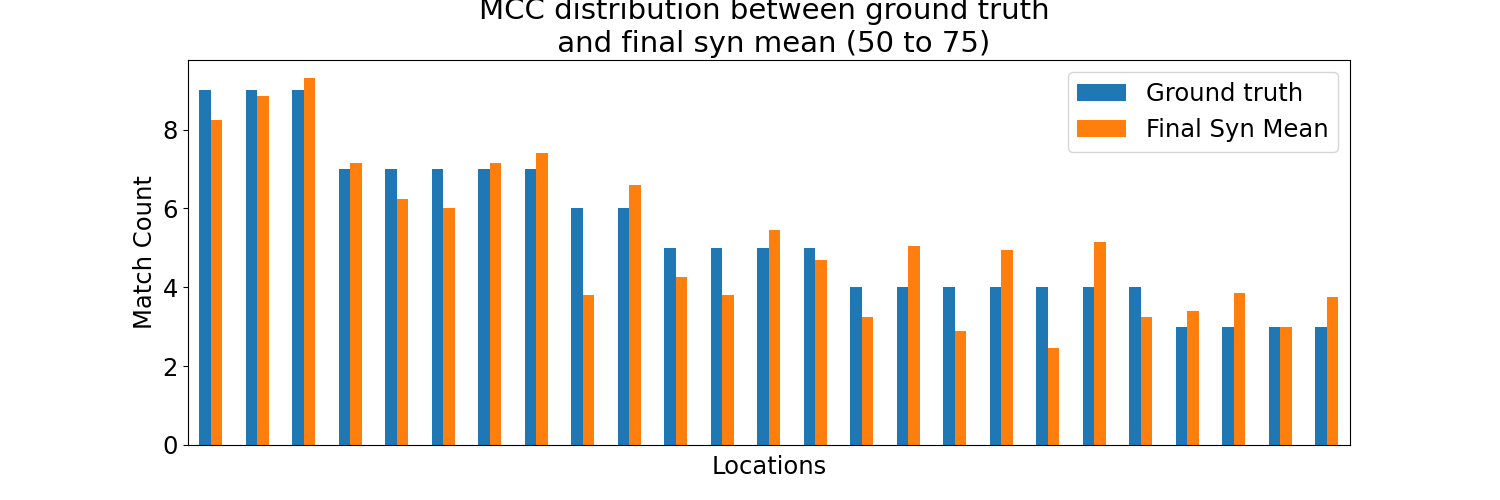}
    \caption{Marginal Distribution of the \texttt{MCC} column. The next 50 entries are shown, and, again, where after the 75th entry the single digit matching counts become insignificant.}
    \label{Appendix:mcc_dist}
\end{figure}

\textbf{Metric-Oriented results.}
\begin{itemize}
    \item Figure \ref{Appendix:time_dependencies} gives the first 300 entries of the relationship between time since each event (transactions) and the transaction amounts. 
    \item Figure \ref{Appendix:fnr_fpr} gives the false negative and false positive rate of the machine learning models trained using encoded synthetic credit card transaction data.
    \item Figure \ref{Appendix:location_dist} gives the 51th to 150th entries of the marginal distribution of \texttt{Location} between original data and synthetic data from Schema 5.
    \item Figure \ref{Appendix:mcc_dist} gives the 26th to 75th entries of the marginal distribution of \texttt{MCC} between original data and synthetic data from Schema 5.
\end{itemize}

\end{document}